\documentclass[runningheads]{llncs}
\usepackage{graphicx}
\usepackage{amsmath}
\usepackage{amssymb}
\usepackage{epstopdf} 
\usepackage{multirow} 
\usepackage{booktabs}
\usepackage{tabularx}
\usepackage{hyperref}
\usepackage{subcaption}
\usepackage{marvosym}

\begin{document}
\title{DCFS: Continual Test-Time Adaptation via Dual Consistency of Feature and Sample}
\titlerunning{DCFS: Continual TTA via Feature and Sample Consistency}
%

\author{Wenting Yin\inst{1} \and
Han Sun\inst{1}\textsuperscript{\Letter} \and
Xinru Meng\inst{1} \and
Ningzhong Liu\inst{1} \and
Huiyu Zhou\inst{2}}
\authorrunning{W. Yin et al.}
\institute{Nanjing University of Aeronautics and Astronautics, Nanjing, China\\
\email{sunhan@nuaa.edu.cn} \and
University of Leicester, UK
}

\maketitle              
\begin{abstract}
Continual test-time adaptation aims to continuously adapt a pre-trained model to a stream of target domain data without accessing source data. Without access to source domain data, the model focuses solely on the feature characteristics of the target data. Relying exclusively on these features can lead to confusion and introduce learning biases. Currently, many existing methods generate pseudo-labels via model predictions. However, the quality of pseudo-labels cannot be guaranteed and the problem of error accumulation must be solved. To address these challenges, we propose DCFS, a novel CTTA framework that introduces dual-path feature consistency and confidence-aware sample learning. This framework disentangles the whole feature representation of the target data into semantic-related feature and domain-related feature using dual classifiers to learn distinct feature representations. By maintaining consistency between the sub-features and the whole feature, the model can comprehensively capture data features from multiple perspectives. Additionally, to ensure that the whole feature information of the target domain samples is not overlooked, we set a adaptive threshold and calculate a confidence score for each sample to carry out loss weighted self-supervised learning, effectively reducing the noise of pseudo-labels and alleviating the problem of error accumulation. The efficacy of our proposed method is validated through extensive experimentation across various datasets, including CIFAR10-C, CIFAR100-C, and ImageNet-C, demonstrating consistent performance in continual test-time adaptation scenarios.
\keywords{Continual Test-Time Adaptation \and Feature Disentanglement Consistency \and Adaptive Loss Weight.}
\end{abstract}
\section{Introduction}
With the wide application of artificial intelligence technology, factors such as data distribution and task requirements faced by models in practical application scenarios may be quite different from their training stages.The continuous test-time adaptation approach requires the model to be continuously updated in an online manner so that it can adapt to new and changing domains.

Prior studies have shown success in handling isolated domain shifts. However, continual adaptation introduces new challenges due to evolving distributions and the absence of source data. One such approach is adjusting batch normalization statistics~\cite{schneider2020improving} during test-time, which can already significantly improve the performance. Another approach updates model weights through self-training methods, such as entropy minimization ~\cite{wang2020tent}. ~\cite{alfarra2023revisiting,niu2022efficient} alleviate error accumulation by filtering out unreliable samples. Robust mean teachers ~\cite{dobler2023robust} use multi-view contrast losses to pull test features toward the original source space and learn about the invariance of the input space.

Motivated by the limitations of current methods in handling entangled features and unreliable pseudo-labels, we propose DCFS, a unified framework that combines feature disentanglement, dual consistency, and confidence-aware learning. The data features of a sample should be composed of two parts: domain features and semantic features. Domain features indicate the difference between two domains, and semantic features contain a common feature representation. Semantic features play an important role in the final classification recognition task. According to the above, the data sample features are disentangled into semantic features and domain features. We fixed the source domain model classifier as a domain-invariant classifier to preserve and learn domain-related information. In addition, an additional semantic classifier is constructed to learn the feature representation of the target data. The pseudo-label of a single sample is fed by the corresponding feature representation into the corresponding classifier to obtain the corresponding prediction output, and then the final more accurate sample prediction probability value is obtained by integrating the two prediction outputs. Finally, the pseudo-label is used for consistency learning with the whole feature output probability value of the sample.

At the same time, in order not to ignore the whole feature information of samples, we introduce self-supervised learning to carry out the whole feature learning of each sample. However, in the presence of domain shift, the model will produce false predicted pseudo-labels for the sample. Setting a higher threshold can ensure the quality of the model predicted pseudo-labels. However, a high threshold will discard many uncertain but possibly correct pseudo-labels, resulting in unbalanced category learning and low pseudo-label utilization.Too low a threshold will inevitably introduce low-quality pseudo labels, which will hurt the performance. Therefore, we introduce an adaptive threshold according to the confident distribution of the model. By calculating a confidence score for each sample to carry out loss weighted learning and enhance the generalization of the target model. Our main contributions can be summarized as follows:

\begin{itemize}
  \item [(1)] 
We perform feature disentanglement on the comprehensive whole features of each sample to distill more meaningful and informative content, enabling the model to acquire independent representations of relevant features.
  \item [(2)]
The dual classifiers are trained to specialize in semantic-related features and domain-related features, respectively, with inherent distinctions. To reinforce their discriminative power, a regularization strategy is employed to maximize the disparity between their parameter sets, thus enhancing the model's ability to differentiate.
  \item [(3)]
To optimize the use of target data, we employ self-supervised learning with loss weighting based on the whole feature of each sample. This approach effectively mitigates the adverse effects of incorrect pseudo-labels generated by the model, reducing the issue of error accumulation.
  \item [(4)]
A large number of experiments on three benchmarks show that the proposed method achieves good performance in multiple benchmarks.
\end{itemize}

\section{Related Work}
\subsection{Unsupervised Domain Adaptation(UDA)}
In these initial UDA frameworks, the focus is primarily on reducing the impact of domain transfer through instance re-weighting techniques~\cite{wang2017instance}.
The objective is to improve the model's performance in the target domain by re-weighting source data instances. Other research focuses on using deep learning techniques to minimize the discrepancy between source and target domain features. This objective was achieved through various methods such as adversarial learning~\cite{ganin2016domain,tzeng2017adversarial}, discrepancy based
loss functions~\cite{chen2020homm,yan2017mind}, or contrastive learning ~\cite{kang2019contrastive,marsden2022contrastive} to achieve.
\subsection{Continual Test-time Adaptation}
During test-time adaptation, the model needs to be adjusted in real time to fit the target data that has distribution shift from the source domain data. Updating the batch normalization (BN) statistics using the target data during test-time shows encouraging results~\cite{schneider2020improving}. 
Both of them do not involve model parameters updates. Most existing methods often update the model weights by self-training in the test time adaptation task. TENT~\cite{wang2020tent} minimizes the entropy of the prediction by updating the BN parameters and performs standard back propagation for model weights updates. Considering the unreliability of high-entropy samples, the loss is only computed based on reliable samples~\cite{niu2022efficient}. Additionally, it uses EWC~\cite{kirkpatrick2017overcoming} regularization to maintain the stability of the model. Similar to EATA~\cite{niu2022efficient}, SAR~\cite{niu2023towards} removes noisy samples with large gradients to ensure that the model weights are optimized to a flat minimum.

Initially, the test-time adaptation approach adapts mainly for a single target domain. ~\cite{wang2022continual} is the first study to consider continual domain changes. It introduces a mean teacher framework to refine pseudo-labels along with stochastic restoration to prevent catastrophic forgetting. One method utilizes contrastive learning and uses symmetric cross-entropy as the consistency loss of mean teacher framework~\cite{dobler2023robust}. Moreover, the different augmented views of one sample are generated to learn transformation-invariant mapping~\cite{choi2022improving}. 
All above methods further reduce error accumulation by introducing self-supervised learning. In AR-TTA~\cite{sojka2023ar}, a small memory combined with mixup data augmentation buffer is used to increase model stability. In contrast to these methods, RoTTA~\cite{yuan2023robust} additionally takes into account the fact that the real-world data is often temporally correlated and proposes a robust approach for this scenario. 
\section{Method}
\begin{figure*}[t!]
    \centering
    \includegraphics[width=12cm]{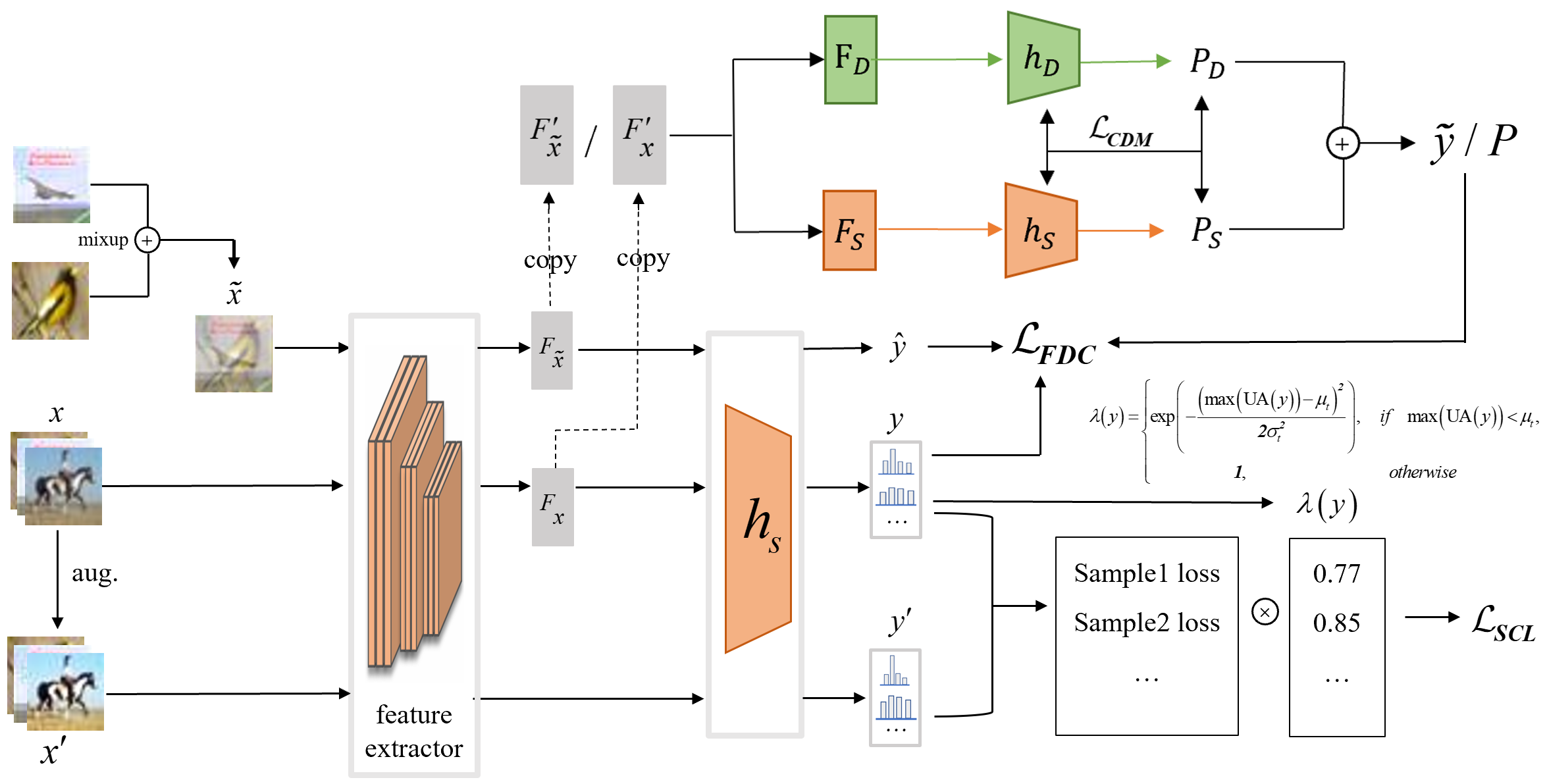}
    \caption{
The framework of our proposed DCFS method.}
    \label{fig:frame}
    \vspace{-0.5cm}
\end{figure*}

The continual test-time adaptation task continuously adapt to the changing target domain in an online manner without using source domain data during test-time, improving the performance of the model pre-trained on the source domain data. The overall architecture of our proposed framework is shown in Fig.\ref{fig:frame}.
\subsection{Feature Disentanglement Consistency}
Among the existing unsupervised domain adaptation methods, most align the entire feature between the two domains in a coarse-grained manner. In this way, not only semantic information, but also domain-related information such as lighting, color, perspective, and environment are aligned. Thus, domain-related information can be treated as noise because it interferes with the feature representation of the classification task. An image should contain both semantic and domain information~\cite{wang2023disentangled}. Semantic information is the determinant of the classification task, while domain information is irrelevant or indistinguishable from the task. A robust model should focus more on domain-invariant semantic information.
\subsubsection{Single Sample Feature Disentanglement Consistency}  Based on the above, the feature of a sample can be intuitively divided into semantic feature and domain feature. For a given target domain image sample $x$, the feature representation $\textbf{F}=g(x)$ can be obtained through the feature extractor of the neural network model. In order to distinguish domain-related feature and semantic-related feature from the whole features, we proposed to use the Coord Attention module $\mathcal{A}$~\cite{hou2021coordinate} to disentangle the features into two parts. Specifically, assume that semantic-related feature is represented as $S$ and domain-related features is represented as $D$. We use $\mathcal{A}$ to generate these two types of feature, which can be expressed as:
\begin{equation}
F_S = \textbf{F} \cdot \mathcal{A}(\textbf{F}), \\
F_D = \textbf{F} \cdot (1- \mathcal{A}(\textbf{F}))
  \label{eq3}
\end{equation}
The domain classifier is represented $h_D(\cdot)$ by a initialized by the source domain model classifier. Domain classifier parameters are kept fixed to maintain domain-related information. In addition, the Feature Disentanglement and Consistency (FDC) module uses the source domain model classifier to initialize an additional semantic classifier representation of the same structure as $h_S(\cdot)$. With the arrival of the target data, the classifier is updated in real time together with the feature extractor to learn the feature representation and semantic information of the target data. By feeding the above two features into two classifiers, two predictive outputs of a sample data under different feature representations can be obtained.
\begin{equation}
P_S = h_S(F_S),   \\
P_D = h_D(F_D)
\end{equation}
Where $P_S$ and $P_D$ are the predicted output under the semantic-related feature and domain-related feature of the sample respectively.

Then, the prediction obtained by the two features are averaged. Finally, we combined different sub-feature prediction to get the implicit whole feature classification result, expressed as $P=P_S+P_D$. The result obtained by the feature disentanglement of the sample should be consistent with its own whole feature prediction. We maintain consistency learning between the prediction of feature disentanglement and the prediction of the whole feature to improve the generalization ability and robustness of the model.
\begin{equation}
\mathcal{L}_{Single(x)} = \mathcal{L}_{CE}(P,y)=-\sum_{c=1}^C P_c \log y_c
\end{equation}
Where $y=h_s(g(x))$ is the predicted output under the whole feature, $C$ represents the number of class of the current task, and $\mathcal{L}_{CE}$ is the cross entropy loss.
\subsubsection{Mixed Sample Feature Disentanglement Consistency}
Through the single sample consistency learning between the feature disentanglement and the whole feature, the false prediction of the model can be improved and the recognition accuracy of the model can be improved. Mixup data augmentation~\cite{zhang2017mixup} mixes two different samples proportionally to create a new training sample and mixes the labels corresponding to the two samples as the label value of the new training sample, to improve the generalization ability of the model.
\begin{equation}
\widetilde x = \rho x_i+(1-\rho)x_j,   \\
\widetilde y = \rho y_i+(1-\rho)y_j,
\end{equation}
Where $\widetilde x$ and $\widetilde y$ respectively represent the mixed target sample and the corresponding label, where $\rho \sim Beta(\alpha,\alpha)$ , for$\alpha \in (0,\infty)$.

In order to make full use of the target data information and make the potential representation of the model more robust for a given task, we use Mixup data enhancement based on feature disentanglement to carry out consistency learning of mixed sample feature disentanglement. Specifically, $y_i$and$y_j$ should have two parts:
\begin{equation}
y_i = P_{iS}+P_{iD},   \\
y_j = P_{jS}+P_{jD}, 
\end{equation}
Thus, the label $\widetilde y$ of each mixed sample can be obtained:
\begin{equation}
\widetilde y = \rho y_i+(1-\rho)y_j= \rho(P_{iS}+P_{iD})+ (1-\rho)P_{jS}+P_{jD}, 
\end{equation}
Similar to the above consistency learning for a single sample, we maintain mixed sample consistency learning between the feature disentanglement result and the whole feature prediction.
\begin{equation}
\mathcal{L}_{Mixup}(\widetilde x) = \mathcal{L}_{CE}(\widetilde y,\hat y)=-\sum_{c=1}^C \widetilde y \log \hat y
\end{equation}
Where $\hat y=h_s(g(\widetilde x))$ is the prediction under the whole feature. The feature disentanglement consistency learning loss is shown as:
\begin{equation}
\mathcal{L}_{FDC} = \mathcal{L}_{Single} + \mathcal{L}_{Mixup}
\end{equation}
\subsection{Classifier Discrepancy Maximization}
We construct double classifiers and use these two different classifiers to realize the learning of different sub-features. Although the two classifiers each make predictions for different features, they should ensure the consistency of the prediction for the same sample. In addition, since the two classifiers learn different feature representations, their internal parameters should show significant differences. This is because each feature representation requires specific parameters to capture and distinguish key information in the data, ensuring that each classifier can work efficiently within its area of expertise. We use $L_1$ regularization term to minimize the classification difference between the two features and maximize the parameter difference between the two classifiers~\cite{wang2023disentangled}.
\begin{equation}
\mathcal{L}_{CDM} =dist(P_S,P_D)+\lambda ||W_1^TW_2||_1
\end{equation}
Where $dist$ represents the similarity measure, and $dist(u,v)=||u,v||_1$. $W_1$ and $W_2$ represent the parameters of the two classifiers, respectively. $\lambda$ is set to the default value 0.1.
\subsection{Sample Consistency Learning}
Sample feature disentanglement splits complex features into simpler sub-features that are easier for the model to process. However, when the feature of the sample is disentangled, it is often too focused on emphasizing the information of individual sub-features, resulting in ignoring the whole feature information of the target domain data. When the model learns domain-related information and semantic-related information, it should avoid losing the understanding and learning of the overall structure of the target domain data. So we feed the whole feature of the data sample into the semantic classifier to get the predictive output. At the same time, an augmented view of the image is generated, and then the image and its augmented view are used to build a self-supervised learning task.

It is biased to use the model's predictive output directly as pseudo-label. Using a fixed threshold for sample filtering can ensure the quality of pseudo-label, but can not balance the quality and quantity of pseudo-label. Based on this, we use a truncated Gaussian function to calculate the confidence score of the sample according to the predicted value of the sample, which can be regarded as a soft version based on threshold. 
Moreover, a Uniform Alignment (UA) method is combined to further solve the problem of unbalanced distribution of pseudo-labels~\cite{chen2023softmatch}. Finally, a weighted cross-entropy loss is constructed between the model's prediction of the augmented data and pseudo-labels for the target data:
\begin{equation}
\mathcal{L}_{SCL} (x)=\lambda (y) \mathcal{L}_{CE} (y,y^{'})
\end{equation}
Where $\lambda (y)$ is the weighted function of the confidence score of $y$ based on the model prediction of the target domain data, and  $y^{'}=h_S(g(x^{'}))$ is the prediction output of the augmented view sample $x^{'}$. We treat the deviation between the confidence maximum $max(y)$ and the Gaussian mean $\mu$ as a proxy measure of the model's prediction correctness. Higher weights are used for pseudo-labels with higher confidence and lower weights are used for more error-prone pseudo-labels with lower confidence.
\begin{equation}
\lambda (y) =
\left \{ 
\begin{array}{cc}
     \lambda_{max} \ exp(-\frac{(max(y)-\mu)^2}{2\sigma^2}), & if \quad max(y)< \mu,  \\
     \lambda_{max} , &  otherwise
\end{array}
\right.
\end{equation}
Where $\lambda_{max}$ is set to 1, the parameters $\mu$ and $\sigma$ in the Gaussian function are unknown. We estimate $\mu$ and $\sigma$ based on each batch prediction of the model for the target domain data.
\begin{equation}
  \begin{split}
     &\mu_b=\mathbb E_B [max(y)]=\frac{1}{B}\sum_{i=1}^B max(y_i) ,\\
&\sigma_b^2=Var_B[max(y)]=\frac{1}{B}\sum_{i=1}^B (max(y_i)-\mu_b)^2
  \end{split}
\end{equation}
$B$ indicates the batch size of the current incoming data. When adapting to the actual test, we used the EMA for both parameters to obtain a more stable estimate.
\begin{equation}
  \begin{split}
     &\mu_t=m \mu_{t-1}+(1-m)\mu_b ,\\
&\sigma_t^2=m \sigma_{t-1}^2+(1-m)\frac{B}{B-1}\sigma_b^2
  \end{split}
\end{equation}
Where $\mu_0$ and $\sigma_0^2$ are initialized to $\frac{1}{C}$ and 1.0, respectively. Since different classes have different learning difficulties, the generated pseudo-label classes may be imbalanced. To solve this problem, we use uniform alignment to generate pseudo-labels with more balanced categories. That is, each prediction on the unlabeled target data is normalized using the ratio between uniform distribution  $\mu(C)$ and $\mathbb E_B [y]$. At the same time, each sample loss weight is calculated using the normalized probability.
\begin{equation}
UA(y)=Normalize(y \cdot \frac{\mu(C)}{\mathbb E_B[y]})
\end{equation}
$Normalize(\cdot)=(\cdot)/\sum(\cdot)$ ensures that the probability of normalization adds up to 1. The final weighting function can be rewritten as:
\begin{equation}
\lambda (y) =
\left \{ 
\begin{array}{cc}
     \lambda_{max} \ exp(-\frac{(max(UA(y))-\mu_t)^2}{2\sigma_t^2}), & if \quad max(UA(y))< \mu_t,  \\
     \lambda_{max} , &  otherwise
\end{array}
\right.
\end{equation}
\subsection{Overall Loss}
In summary, the overall loss of the DCFS model is as follows:
\begin{equation}
\mathcal{L}=\mathcal{L}_{FDC}+\lambda_{CDM} \mathcal{L}_{CDM}+\lambda_{SCL} \mathcal{L}_{SCL}
\end{equation}
Where $\lambda_{CDM}$ and $\lambda_{SCL}$ are the tradeoff factors for balancing losses.
\section{Experiments}
\subsection{Experimental Setup}
 \subsubsection{Datasets}
We illustrate the effectiveness of our method on three commonly benchmark datasets, including ImageNet-C, CIFAR10-C, and CIFAR100-C. These datasets include 15 different types of image corruption with 5 levels of severity. We use a continual benchmark~\cite{wang2022continual}. The model is adapted to a sequence of test domains in an online manner.
 \subsubsection{Implementation Details}
In all experiments, we use WideResNet-28~\cite{zagoruyko2016wide}, ResNeXt-29~\cite{xie2017aggregated} and the uniformly pre-trained ResNet50 model in the CIFAR10 to CIFAR10-C, CIFAR100 to CIFAR100-C, and ImageNet to ImageNet-C tasks, respectively, all of which are taken from the RobustBench benchmark~\cite{croce2020robustbench}. For CIFAR10-C and CIFAR100-C datasets, we use Adam optimizer with a learning rate of 1e-3 and the batch size is set to 200. For ImageNet-C dataset, we use the SGD optimizer with a learning rate of 0.0002 and the batch size is set to 32. 
Regarding the hyperparameter settings of our method, both $\lambda_{CDM}$ and $\lambda_{SCL}$ are set to 1.0.
\subsubsection{Baselines}
We compared our method with several state-of-the-art methods, including source-only: Source, BN
Adapt~\cite{schneider2020improving}(NIPS 2020), TENT~\cite{wang2020tent}(ICLR 2021),
CoTTA~\cite{wang2022continual}(CVPR 2022), RoTTA~\cite{yuan2023robust}(CVPR 2023), AR-TTA~\cite{sojka2023ar}(ICCV 2023),
ECoTTA~\cite{song2023ecotta}(CVPR 2023), LAW~\cite{park2024layer}(WACV 2024), PALM~\cite{maharana2024palm}(AAAI 2025).

\begin{table*}[ht!]
\centering
\caption{Classification error rate~(\%) for the CIFAR10-to-CIFAR10-C continual test-time adaptation task. Results are evaluated on the largest corruption severity level 5. }\label{tab:cifar10}
\vspace{-2.5mm}
\scalebox{0.76}{
\tabcolsep3pt
\begin{tabular}{l|c c c c c c c c c c c c c c c|c}
\multicolumn{1}{l}{}& \multicolumn{15}{l}{$t\xrightarrow{\hspace*{11.5cm}}$}& \\ \hline
 method & \rotatebox[origin=c]{70}{gaussian} & \rotatebox[origin=c]{70}{shot} 
 & \rotatebox[origin=c]{70}{impulse} & \rotatebox[origin=c]{70}{defocus} 
 & \rotatebox[origin=c]{70}{glass} & \rotatebox[origin=c]{70}{motion} 
 & \rotatebox[origin=c]{70}{zoom} & \rotatebox[origin=c]{70}{snow} 
 & \rotatebox[origin=c]{70}{frost} & \rotatebox[origin=c]{70}{fog} 
 & \rotatebox[origin=c]{70}{brightness} & \rotatebox[origin=c]{70}{contrast} 
 & \rotatebox[origin=c]{70}{elastic} & \rotatebox[origin=c]{70}{pixelate}
 & \rotatebox[origin=c]{70}{jpeg} & Mean\\
\hline
Source & 72.3  & 65.7 & 72.9 & 46.9 & 54.3  & 34.8   & 42.0 & 25.1 & 41.3   & 26.0 & 9.3  & 46.7 & 26.6  & 58.5 & 30.3  & 43.5 \\

 BN Adapt & 28.1  & 26.1  & 36.3  & 12.8  & 35.3  & 14.2  & 12.1  & 17.3 & 17.4  & 15.3 & 8.4   & 12.6  & 23.8  & 19.7   & 27.3 & 20.4 \\

 TENT-cont.  & 24.8 &20.6 &	28.6 &	14.4 &	31.1 &	16.5 &	14.1 &	19.1 &	18.6 &	18.6 &	12.2 &	20.3 &	25.7 &	20.8 &	24.9 &	20.7\\

 CoTTA & 24.3  & 21.3 & 26.6  & 11.6  & 27.6  & 12.2 & 10.3 & 14.8 & 14.1  
 & 12.4 & 7.5 & 10.6    & 18.3 & 13.4  & 17.3 & 16.2 \\
 RoTTA &30.3 &25.4 &34.6 &18.3 &34.0 &14.7 &11.0 &16.4 &14.6 &14.0 &8.0 &12.4 &20.3 &16.8 &19.4 &19.3 \\
 AR-TTA &30.8 &25.2 &33.6 &15.5 &32.2 &16.3 &14.8 &18.6 &17.3 &16.6 &12.0 &15.3 &26.1 &21.4 &23.0 &21.2\\
 ECoTTA &23.8 &18.7 &25.7 &11.5 &29.8 &13.3 &11.3 &15.3 &15.0 &13.0 &7.9 &11.3 &20.2 &15.1 &20.5 &16.8\\
 LAW &24.7 &18.9 &25.5 &12.9& 26.7 &15.0 &11.8 &15.1 &14.7 &15.8 &10.1 &13.8 &19.4 &14.7 &18.3 &17.2\\
  PALM &25.8 &\textbf{18.1} &\textbf{22.7} &\textbf{12.3} &\textbf{25.3} &13.1 &10.7 &\textbf{13.5} &13.1&12.2 &8.5 &11.8 &\textbf{17.9} &\textbf{12.0} &\textbf{15.4} &\textbf{15.5}\\
 \hline
  ours&\textbf{23.3}  &19.7  &24.3  &\textbf{12.3} &26.1  &\textbf{12.1}  &\textbf{9.9}  &14.2 &\textbf{12.2}  &\textbf{11.5}  &\textbf{6.8} & \textbf{10.2}  &18.9 &13.6 &17.8 &\textbf{15.5} \\

\hline
\end{tabular}} 
\vspace{-0.25cm}
\end{table*}

\begin{table*}[ht!]
\centering
\caption{Classification error rate~(\%) for the CIFAR100-to-CIFAR100-C continual test-time adaptation task. Results are evaluated on the largest corruption severity level 5. }\label{tab:cifar100}
\vspace{-2.5mm}
\scalebox{0.76}{
\tabcolsep3pt
\begin{tabular}{l|c c c c c c c c c c c c c c c|c}
\multicolumn{1}{l}{}& \multicolumn{15}{l}{$t\xrightarrow{\hspace*{11.5cm}}$}& \\ \hline
 method & \rotatebox[origin=c]{70}{gaussian} & \rotatebox[origin=c]{70}{shot} 
 & \rotatebox[origin=c]{70}{impulse} & \rotatebox[origin=c]{70}{defocus} 
 & \rotatebox[origin=c]{70}{glass} & \rotatebox[origin=c]{70}{motion} 
 & \rotatebox[origin=c]{70}{zoom} & \rotatebox[origin=c]{70}{snow} 
 & \rotatebox[origin=c]{70}{frost} & \rotatebox[origin=c]{70}{fog} 
 & \rotatebox[origin=c]{70}{brightness} & \rotatebox[origin=c]{70}{contrast} 
 & \rotatebox[origin=c]{70}{elastic} & \rotatebox[origin=c]{70}{pixelate}
 & \rotatebox[origin=c]{70}{jpeg} & Mean\\
\hline
Source& 73.0&	68.0&	39.4&	29.3&	54.1&	30.8&	28.8&	39.5&	45.8&	50.3&	29.5&	55.1&	37.2	&74.7&	41.2&	46.4\\
BN Adapt & 42.1     & 40.7 & 42.7    & 27.6    & 41.9  & 29.7   & 27.9 & 34.9 & 35.0   & 41.5 & 26.5       & 30.3     & 35.7           & 32.9     & 41.2 & 35.4 \\

TENT-cont.  & 37.2     & 35.8 & 41.7    & 37.9    & 51.2  & 48.3   & 48.5 & 58.4 & 63.7   & 71.1 & 70.4       & 82.3     & 88.0           & 88.5     & 90.4 & 60.9 \\

CoTTA     & 40.1  &	37.7   &39.7  &26.9  &38.0   &27.9   &26.4   &32.8&	31.8   &40.3     &	24.7&	26.9&	32.5   &28.3&	33.5   &32.5\\
RoTTA  &49.1 &44.9 &45.5 &30.2 &42.7 &29.5 &26.1 &32.2 &30.7 &37.5 &24.7 &29.1 &32.6 &30.4 &36.7 &34.8 \\
 LAW &41.0 &36.7 &38.3 &25.6 &37.0 &27.8 &25.2 &30.7 &30.0 &37.3 &24.4 &27.6 &31.2 &27.8 &34.9 &31.7\\
 PALM &\textbf{37.3}& \textbf{32.5} &34.9 &26.2& \textbf{35.2} &27.5 &24.6 &\textbf{28.8} &\textbf{29.1} &\textbf{34.1} &23.5 &26.9 &31.1 &26.5 &34.1 &30.1
 \\
\hline
ours &37.5 &35.0 & \textbf{33.5} &\textbf{24.7} &36.3 &\textbf{25.8} &\textbf{24.5} &30.4 &29.2 &35.0 &\textbf{23.2} &\textbf{25.1} &\textbf{31.0} &\textbf{26.4} &\textbf{33.4} &\textbf{30.0 }\\
\hline
\end{tabular}} 
\vspace{-0.25cm}
\end{table*}

\begin{table*}[ht!]
\centering
\caption{Classification error rate~(\%) for the ImageNet-to-ImageNet-C continual test-time adaptation task. Results are evaluated on the largest corruption severity level 5.}\label{tab:imagenet}
\vspace{-2.5mm}
\scalebox{0.76}{
\tabcolsep3pt
\begin{tabular}{l|c c c c c c c c c c c c c c c|c}
\multicolumn{1}{l}{}& \multicolumn{15}{l}{$t\xrightarrow{\hspace*{11.5cm}}$}& \\ \hline
 method & \rotatebox[origin=c]{70}{gaussian} & \rotatebox[origin=c]{70}{shot} 
 & \rotatebox[origin=c]{70}{impulse} & \rotatebox[origin=c]{70}{defocus} 
 & \rotatebox[origin=c]{70}{glass} & \rotatebox[origin=c]{70}{motion} 
 & \rotatebox[origin=c]{70}{zoom} & \rotatebox[origin=c]{70}{snow} 
 & \rotatebox[origin=c]{70}{frost} & \rotatebox[origin=c]{70}{fog} 
 & \rotatebox[origin=c]{70}{brightness} & \rotatebox[origin=c]{70}{contrast} 
 & \rotatebox[origin=c]{70}{elastic} & \rotatebox[origin=c]{70}{pixelate}
 & \rotatebox[origin=c]{70}{jpeg} & Mean\\
\hline

Source  & 95.3 & 94.5 & 95.3 & 84.8 & 91.0 & 86.8 & 77.1 & 84.3 & 79.7 & 77.2 & 44.4 & 95.5 & 85.2 & 76.9 & 66.6 & 82.3 \\
BN Adapt   & 88.0 & 88.0 & 88.1 & 88.9 & 87.6 & 78.7 & 66.4 & 68.5 & 71.0 & 56.3 & 37.3 & 89.8 & 59.8 & 57.7 & 68.0 & 72.9 \\
TENT-cont. & 84.2 & 78.0 & 76.4& 81.9 & 80.4 & 74.2 & 64.0 & 70.6 & 71.8 & 62.9 & 48.3 &86.0 &65.2 & 61.1 & 68.4 & 71.6 \\
CoTTA    & 87.0 & 82.9 & 78.2 & 81.8 & 78.4 & 71.0 & 64.3 & 67.4 & 67.1 & 60.6 & 53.8 & 70.3 & 60.3 & 57.1 & 60.1 & 69.3 \\
RoTTA   &88.4 &83.0 &82.1 &91.5 &83.5 &72.9 &59.9 &67.5 &64.6 &53.9 &35.1 &74.5 &54.5 &48.4 &52.9 &67.5\\
AR-TTA  &-&-&-&-&-  &-&-&-&-&-  &-&-&-&-&- &68.0\\
 ECoTTA&-&-&-&-&-  &-&-&-&-&-  &-&-&-&-&- &63.4\\
  LAW &81.8 &75.0 &72.1 &77.4 &73.2 &63.9 &54.6 &57.9 &61.0 &50.1 &36.3 &68.6 &49.08 &46.2 &49.1 &61.1\\
 PALM &81.0 &73.3& 70.8 &77.0 &\textbf{71.8} &62.3 &53.9 &56.7 &\textbf{60.7} &50.3 &36.2 &\textbf{65.9}&\textbf{48.0} &\textbf{45.2} &\textbf{48.0}&60.1\\
\hline
ours &\textbf{77.6} & \textbf{72.4} &\textbf{70.8} &\textbf{73.8} &74.7 &\textbf{61.9} &\textbf{51.5} &\textbf{55.9} &61.3 &\textbf{46.2} &\textbf{35.6} &70.8 &51.3 &45.7 &48.9 &\textbf{59.8} \\
\hline
\end{tabular}} 
\vspace{-0.5cm}
\end{table*}
\subsection{Results and Analysis}
The experimental results on CIFAR10-C, CIFAR100-C, and ImageNet-C are summarized in Tables 1–3. Table ~\ref{tab:cifar10} shows the performance of DCFS on the CIFAR10-C task. DCFS achieved the lowest average error rate and was $0.7\%$ lower than CoTTA that is the next lowest error rate method. On the CIFAR100-C dataset, DCFS achieved optimal performance on 15 domain adaptation tasks. Whether faced with blur, noise, contrast changes, or other types of image degradation, DCFS can cope effectively. ImageNet-C data set is a data set with large data, many classification tasks and difficult classification. When the domain adaptation task is challenged using only the source domain model, the error rate on the $Gau.$ task is as high as $95.3\%$, and the recognition performance on some subsequent tasks such as $Imp.$ and $Con.$ is extremely poor. The features of data samples are divided into semantic features and domain features. Focus on learning domain-related features while keeping the source domain model's classifier fixed. Additional semantic classifiers are introduced to capture relevant features of the target data. The average prediction of sub-features and the consistency learning of the whole features enhance the generalization ability and robustness of the model. We employ loss-weighted learning for each sample according to its confidence to further improve the utilization efficiency of the model for weak learning class. In the three tasks, the performance of DCFS model has been greatly improved compared with the method without domain adaptation technology and the existing advanced methods, which confirms the effectiveness of DCFS.
\subsection{Ablation Studies}
\subsubsection{Component Analysis} 
In order to confirm the role of the proposed modules in the adaptation phase, the results of the ablation experiments on the effects of different modules on the model performance are shown in Table~\ref{tab:tab4}. The first row of the table shows the results using only the source model as the baseline for comparison, at which point the model has a classification error rate of $43.5\%$. When disentanglement of the feature of the target data sample feature disentanglement and consistency learning are performed, the error rate is reduced by $8.9\%$ compared to using only the source model, indicating that the learning of the target features becomes more explicit. Then, by introducing the regularization method to maximize the parameter difference between the two classifiers, we enhance the model differentiation ability and further improve the performance of the model. Finally, sample consistency learning is carried out on the feature of target data sample, makes the model on the features of target data more comprehensive study. When we combine the feature disentanglement of the target data with its whole feature, the generalization ability of the model is greatly improved, and the average error rate is directly reduced to $19.3\%$. The three modules work together to achieve the best performance.
\subsubsection{Hyperparameters} 
\begin{figure}
\centering
  \vspace{-0.2cm}
\begin{tabular}{cc}    
\includegraphics[width=6cm]{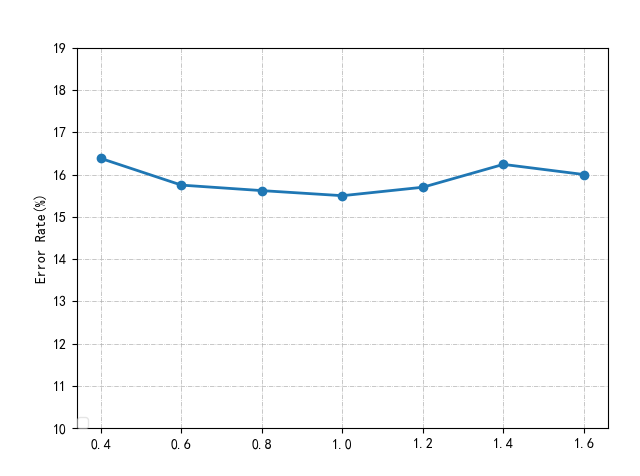} &
\includegraphics[width=6cm]{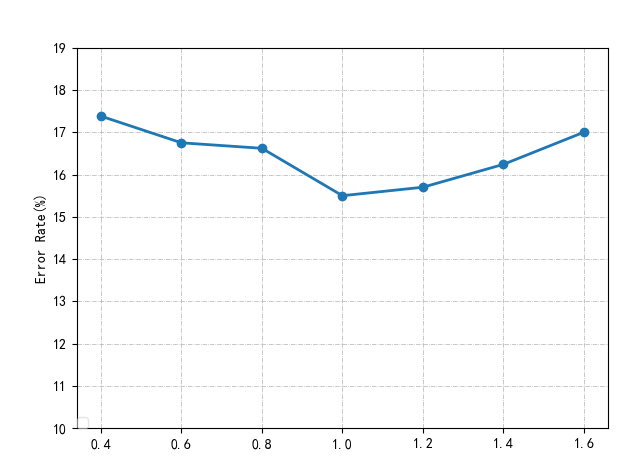}\\
\vspace{-0.6cm}
\end{tabular}
\caption{ Results of parameter $\lambda_{CDM}$(left) and $\lambda_{SCL}$(right) sensitivity.}
\label{fig:fig2}
 \vspace{-0.55cm}
\end{figure}

In this subsection, we evaluate the sensitivity of hyperparameters with the tasks CIFAR10-to-CIFAR10-C. Fig.~\ref{fig:fig2} shows the sensitivity of performance for values of $\lambda_{CDM}$ of [0.4,0.6,0.8,1.0,1.2,1.4,1.6] and values of $\lambda_{SCL}$ of [0.4,0.6,0.8,1.0,1.2,1.4,1.6]. The results show that our method is insensitive to both parameters.

\begin{table}
    \centering
    \vspace{0.1cm}
    \caption{Ablation of the losses on CIFAR10-C.}
    \begin{tabular}{ccc ccc ccc ccc|ccc}
    \hline
    Backbone(Source)&&  &$\mathcal{L}_{FDC}$&&   &$\mathcal{L}_{CDM}$&& &$\mathcal{L}_{SCL}$&&     &&Avg.& \\    
    \hline
    \checkmark&&  &&&   &&&  &&&      &&43.5& \\
    \checkmark&&  &\checkmark&& &&& &&& &&34.6& \\
    \checkmark&&  &\checkmark&&   &\checkmark&&   && &         &&28.5& \\
    \checkmark&&  &\checkmark&&   &&&  
    &\checkmark&&  &&19.3& \\
    \checkmark&&  &\checkmark&&   &\checkmark&&   &\checkmark&&  &&15.5& \\
    \hline
    \end{tabular}
    \label{tab:tab4}
      \vspace{-0.19cm}
\end{table}

\section{Conclusion}
We propose a new framework called DCFS. By disentangling the whole feature of the sample, the model can learn the representation of each feature independently, which helps to extract more essential and useful information. The regularization is used to maximize the parameter difference between the two classifiers 
to enhance the ability of the model to distinguish different features. Finally, we perform self-supervised loss-weight for the whole predicted value of the sample to alleviate the negative impact of false pseudo-labels.


%
%
%
\bibliographystyle{splncs04}
\bibliography{ref}
%





\end{document}